\documentclass{article}

 \usepackage[preprint]{sty_mod}

\usepackage[utf8]{inputenc} %
\usepackage[T1]{fontenc}    %
\usepackage{hyperref}       %
\usepackage{url}            %
\usepackage{booktabs}       %
\usepackage{amsfonts}       %
\usepackage{nicefrac}       %
\usepackage{microtype}      %
\usepackage{xcolor}         %

\usepackage{graphicx}
\usepackage{caption}
\usepackage{subcaption}
\usepackage{amsmath}
\usepackage{cleveref}
\usepackage{booktabs}
\usepackage{xcolor}
\usepackage{colortbl}
\usepackage{graphicx}
\usepackage[most]{tcolorbox}
\newcommand{\given}{\ |\ }

\definecolor{takeawayfill}{HTML}{FFF1E6}
\definecolor{takeawayframe}{HTML}{E0833C}
\newtcolorbox{takeaway}{
  enhanced,
  colback=takeawayfill,
  colframe=takeawayframe,
  boxrule=0.8pt,
  arc=4pt,
  left=8pt, right=8pt, top=6pt, bottom=6pt,
  fonttitle=\bfseries,
  coltitle=takeawayframe,
  title=Takeaway,
  attach title to upper=\ ,
  after title=.\ ,
}

\definecolor{pavelgreen}{RGB}{0,170,60}
\definecolor{pinkypink}{RGB}{239, 71, 111}
\ifdefined\hidetodos
  \newcommand{\xxcomment}[4]{}
  \newcommand{\vatsal}[1]{}
  \newcommand{\pavel}[1]{}
  \newcommand{\andrew}[1]{}
  \newcommand{\shikai}[1]{}
  \newcommand{\charlie}[1]{}
\else
  \newcommand{\xxcomment}[4]{\textcolor{#1}{[$^{\textsc{#2}}_{\textsc{#3}}$ #4]}}
  \newcommand{\vatsal}[1]{\xxcomment{pinkypink}{V}{B}{#1}}
  \newcommand{\pavel}[1]{\xxcomment{pavelgreen}{P}{I}{#1}}
  
  \newcommand{\shikai}[1]{\xxcomment{orange}{S}{Q}{#1}}
  \newcommand{\charlie}[1]{\xxcomment{purple}{C}{C}{#1}}
\fi

\title{Emergent Capabilities Arise Randomly from Learning Sparse Attention Patterns}

\author{%
  Vatsal Baherwani \\
  New York University \\
  \texttt{vatsalbaherwani@nyu.edu} \\
  \And
  Zixi Chen \\
  New York University
  \And
  Shikai Qiu \\
  New York University
  \AND
  Andrew Gordon Wilson \\
  New York University
  \And
  Pavel Izmailov \\
  New York University\\
}

\begin{document}

\maketitle

\begin{abstract}
Neural scaling laws for transformer language models predict smooth improvements in pretraining loss with increasing parameters, but downstream capabilities such as in-context learning are known to emerge abruptly past a certain model scale. In this paper, we show that emergent capabilities arise stochastically throughout training, with larger models acquiring them earlier on average. We demonstrate that the emergence of capabilities such as pattern completion and indirect object identification corresponds to the abrupt learning of task-relevant attention patterns. To isolate this phenomenon, we train transformer models on synthetic linear map and cellular automata datasets, and we show that the difficulty of learning attention patterns depends on context length and pattern sparsity.  Moreover, scaling the number of attention heads improves learning efficiency on our synthetic tasks, while increasing the head dimension yields diminishing returns past a minimum capacity. We additionally investigate architectures with alternative attention mechanisms, showing that MLP-Mixer outperforms a transformer on linear map tasks with complex attention patterns. Our findings provide a mechanistic insight into emergence, showing that downstream capabilities arise abruptly due to the intrinsic difficulty of learning sparse attention patterns in transformer models. 
\end{abstract}

\section{Introduction}
As transformer language models grow in parameter count, they acquire emergent capabilities not seen in smaller models~\citep{wei2022emergentabilitieslargelanguage}; however, it remains unclear what changes inside a model when a capability emerges. At a fixed scale, whether a capability appears by the end of training depends on the initialization seed~\citep{zhao2026randomscalingemergentcapabilities}, and skills such as syntax acquisition and in-context learning emerge abruptly within a short window of training~\citep{chen2025suddendropslosssyntax, olsson2022incontextlearninginductionheads}. Are there interpretable mechanisms behind emergence?

\begin{figure}[tbp]
    \centering
    \includegraphics[width=0.8\linewidth]{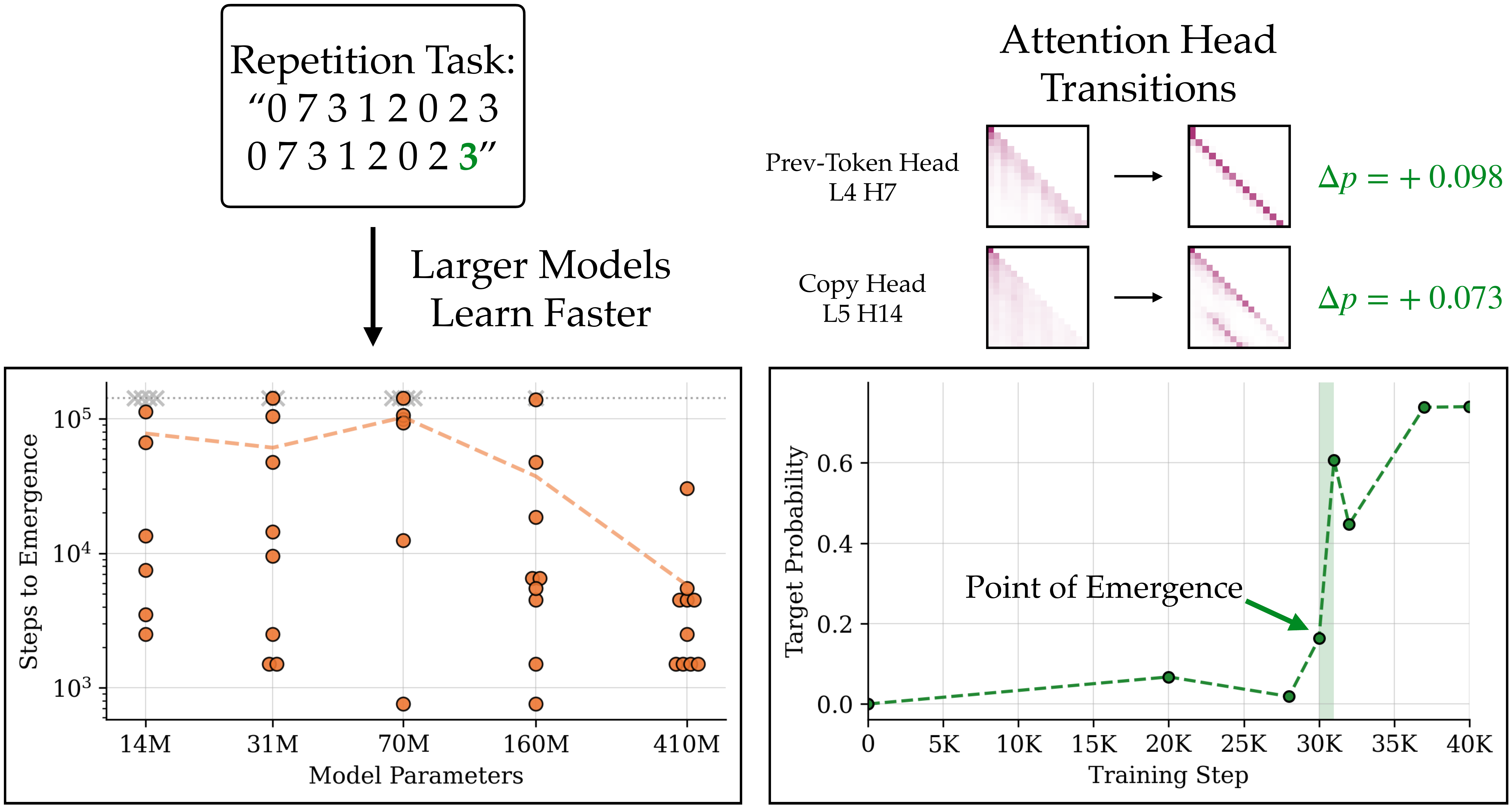}
    \caption{\textbf{Language model capabilities emerge randomly and abruptly due to learning core attention patterns.} Left: models with different initialization seeds learn a repetition task at random points throughout training; larger models consistently learn faster on average (dashed line). Models marked with a grey X did not solve the task by the end of training. Right: the model's correct token probability abruptly spikes during training, which coincides with learning relevant attention patterns.}
    \label{fig:emergence}
\end{figure}

In this paper, we show that the \textbf{emergence of capabilities} in language models is driven by \textbf{learning task-relevant attention patterns}. At a fixed model scale, a capability may arise at variable points during training or even fail to arise depending on the model initialization. When emergence does occur, it coincides with the acquisition of task-relevant attention patterns, often across multiple heads~(Figure~\ref{fig:emergence}). Larger models acquire these attention patterns earlier and more reliably across initializations, which we argue is the mechanism behind the higher emergence rates reported by~\citet{zhao2026randomscalingemergentcapabilities}. We also demonstrate that attention learning is the key bottleneck to emergent capabilities~(Section~\ref{sec:emergence}), as patching learned attention maps into an earlier training checkpoint can recover most of the performance for a given capability.

In natural language, there is no single ground-truth attention pattern, making it impossible to systematically isolate how factors such as context length and sparsity affect the difficulty of learning attention patterns. We study these factors in a controlled setting by training transformer models on synthetic linear map and cellular automata datasets 
(Section~\ref{sec:synthetic}). Notably, we find that varying the context length or attention pattern sparsity dictates whether a model solves a task completely or makes no progress at all. We then investigate the role of architectures in learning attention patterns in Section~\ref{sec:arch}. Scaling width yields consistent benefits on both of our attention tasks, particularly when increasing the number of heads given a sufficiently high head dimension. We also investigate alternative architectures and find that MLP-Mixer~\citep{tolstikhin2021mlpmixerallmlparchitecturevision}, which uses an MLP for its token-mixing mechanism rather than dot-product attention, outperforms a transformer by an order of magnitude in learning the ground-truth attention pattern for our linear map task.
Although we only observe this improvement on a single synthetic task, it suggests a tantalizing possibility of designing architectures with more sample-efficient token-mixing mechanisms than transformers. 

Through our emergence analysis on Pythia models and training experiments on synthetic tasks, we observe that learning long-context, sparse attention patterns is a key bottleneck for eliciting downstream capabilities in transformer models. 
We see that various architectural choices such as scaling attention heads or applying entirely new token-mixing mechanisms can significantly improve a model's ability to learn these patterns. Our findings prescribe multiple future research directions on designing datasets and architectures to improve a model's ability to learn sparse attention patterns such as copying or induction heads, thereby removing a key bottleneck in the emergence of capabilities and enabling more efficient scaling of downstream performance.

\section{Random Emergence of Capabilities in Language Models}
\label{sec:emergence}

While neural scaling laws~\citep{kaplan2020scalinglawsneurallanguage, hestness2017deeplearningscalingpredictable} predict a smooth decrease in training loss as a function of model size, certain downstream capabilities are known to emerge both \textit{abruptly} at certain model scales~\citep{wei2022emergentabilitieslargelanguage} and \textit{randomly} depending on the model initialization~\citep{zhao2026randomscalingemergentcapabilities}. Specifically, \citet{zhao2026randomscalingemergentcapabilities} demonstrate that model performance on many downstream tasks follows a bimodal distribution: different models at a fixed scale will either reliably solve a task or fail entirely depending on the random initialization seed, with the success probability increasing at larger model scales. This observation suggests that although language model training dynamics are smooth across training steps, the acquisition of certain capabilities may also occur randomly throughout training. We confirm this hypothesis and demonstrate that sharp, random emergence of capabilities during training occurs when the model learns core attention patterns relevant to a certain task.

All experiments in this section are conducted with the Pythia suite of language models~\citep{biderman2023pythiasuiteanalyzinglarge}. Each set of Pythia checkpoints contains model weights at initialization, at each step corresponding to powers of 2 between $1$ and $512$, and then at every $1000$ steps up until the final checkpoint at step $143000$. We study the models with 14M, 31M, 70M, 160M, and 410M parameters, with ten sets of checkpoints for each model scale corresponding to different random initializations.

\subsection{Searching for Emergence During Training}
\label{sec:emergence_search}

We first evaluate how often Pythia models can solve a given task by the end of training. We construct a task based on a single sample, where the model is provided with a prefix and must predict the correct next token. Our analysis focuses on single-sample evaluations due to slight variations in emergence times when introducing multiple samples. We provide additional details on our motivation behind single-sample evaluations and confirm our findings apply to multi-sample evaluations in Appendix~\ref{app:sample}.
We define a capability as having emerged for a given model size and initialization seed if the most likely next token (i.e., the model output under greedy sampling) matches the ground-truth output at the final training checkpoint. Formally, emergence occurs when $\arg\max_{y\in \mathcal V}p(y\ |\ x)=y^*$ given a model with token vocabulary $\mathcal V$ where $y^*$ is the ground-truth output given prefix $x$. For the repetition task in Figure~\ref{fig:emergence}, emergence corresponds to the model predicting ``9'' given the prefix ``1 \dots 9 1 \dots 8''.

If a model does learn to solve a given task, we search for the point during training at which the capability first emerges. We define the \textit{point of emergence} as the smallest $t$ such that $\arg \max_{x\in \mathcal V} p_t(y\given x)=y^*$, where $p_t(x)$ is the model prediction at training step $t$. We assume that after the point of emergence, the model will not revert to another incorrect prediction at a later timestep $t'>t$, and we verify this claim holds empirically. With this assumption, we apply binary search (rather than evaluating all 154 checkpoints) to find the first checkpoint where $\arg \max_{x\in \mathcal V} p_t(y\given x)=y^*$. Because we can only evaluate $p_t(y\given x)$ at specific checkpoints, we estimate $t$ as the midpoint of the first checkpoint step with emergence and the checkpoint prior to it. We apply this approach to compute the point of emergence for all ten model seeds at each scale. Figure~\ref{fig:emergence} (left) illustrates the results. 

At smaller model scales, some models successfully learn the repetition task, but for specific initialization seeds the capability never emerges. As previously demonstrated by~\citet{zhao2026randomscalingemergentcapabilities}, we see that the rate at which models acquire a capability increases with scale; at the 160M and 410M parameter scales, the task is solved by all ten models. Notably, we also see that the same capability can emerge randomly at different points throughout training, and larger models acquire capabilities earlier on average.

\subsection{Learning Core Attention Patterns Enables Abrupt Emergence}
In Figure~\ref{fig:emergence} (right) we plot $p_t(y^*\given x)$, the next-token probability the model assigns to the ground-truth answer, as a function of training steps $t$. We see that $p_t(y^*\given x)$ remains very small up until the point of emergence, and then abruptly jumps between the point of emergence and the previous training checkpoint. Given that emergence occurs abruptly, we investigate how a language model's internal representations change before and after acquiring this capability.

\paragraph{Causal Attention Head Ablation.} When training transformers on algorithmic tasks, there is often a fixed ground-truth attention pattern the model must learn. \citet{gopalani2025happenslossplateauunderstanding}~show that learning this ground-truth attention map is a major bottleneck that leads to abrupt learning, and biasing the model's attention map toward this optimal solution significantly accelerates training convergence. While there is no single ground-truth attention pattern for next-token prediction on natural language, we hypothesize that the abrupt learning we observe on individual capabilities is also due to sudden changes in the model's attention weights. We validate this hypothesis by tracking how each attention head in the language model influences the sudden increase in $p_t(y^*\given x)$. We consider the first checkpoint that exhibits emergence as the \textit{post-emergence} model, and the preceding checkpoint as the \textit{pre-emergence} model. For each attention head at every layer, we observe the head's attention map computed on the prefix $x$ in the post-emergence model. We then input the same prefix to the pre-emergence model, while fixing the attention pattern for that head to be that of the post-emergence model. This approach is effectively activation patching~\citep{heimersheim2024useinterpretactivationpatching} on the dot-product attention scores. We now test whether the attention patterns alone can explain the sudden boost in model capability.

We record the change in $p_t(y^*\given x)$ resulting from patching each individual head in the pre-emergence model, and select the top $K=16$ heads with the largest change $\Delta p$. We refer to these top $K$ heads as \textit{causal attention heads}. Figure~\ref{fig:emergence} (top right) visualizes the change in attention patterns of two such heads. We see both attention heads transition from a roughly uniform map to interpretable previous-token and copy patterns, which together enable the model to solve the repetition task. As we show in Figure~\ref{fig:emergence_tasks}, \textbf{patching the attention heads alone recovers the performance gap} between the pre-emergence and post-emergence models. The effectiveness of patching depends on our choice of $K$; rather than attributing the rapid increase in $p_t(y^*\given x)$ to a single head, we find that \textbf{emergence jointly depends on multiple attention heads} learning complementary patterns. For example, in Figure~\ref{fig:emergence} both the previous-token and copy head are necessary to solve the repetition task. We provide additional details regarding our causal ablation implementation in Appendix~\ref{app:causal}.

\subsection{Emergence Across Multiple Tasks}
\begin{figure}[t]
    \centering
    \includegraphics[width=1\linewidth]{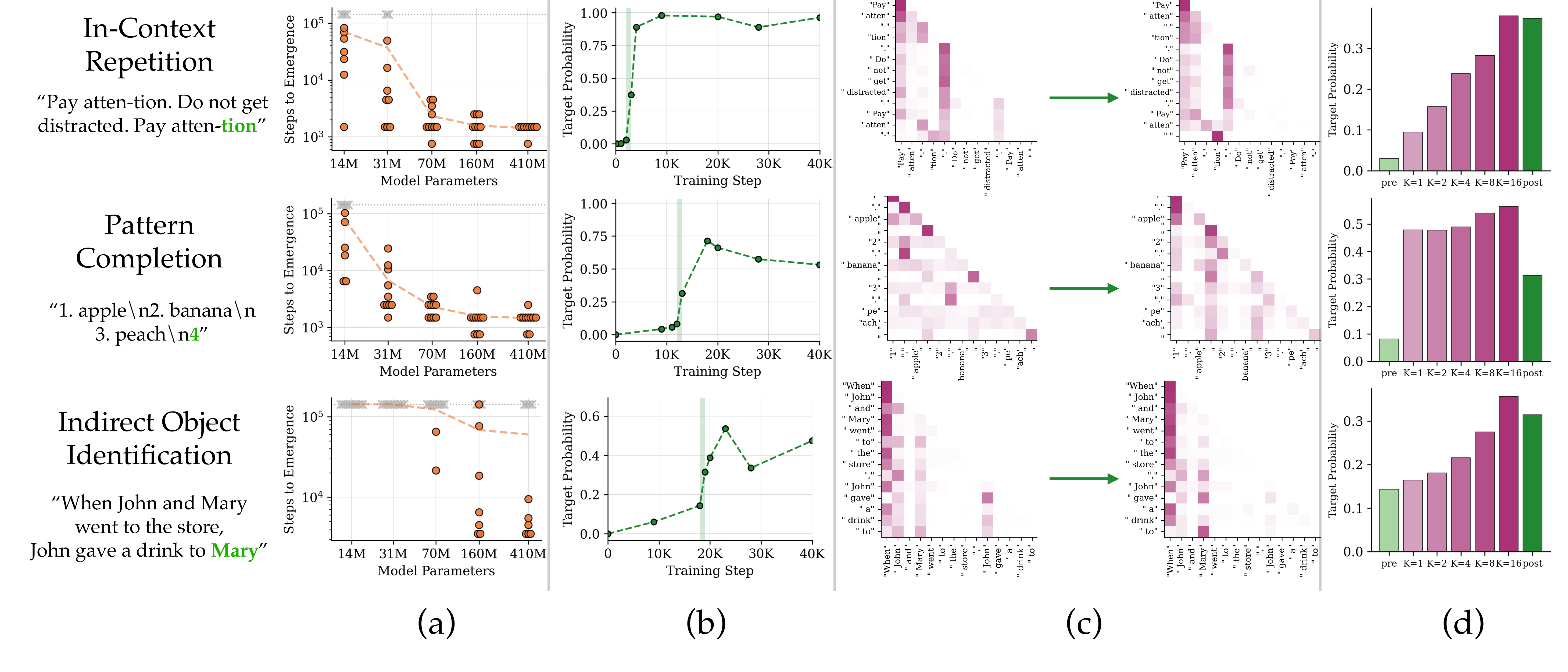}
    \caption{
      \textbf{Random emergence across three tasks.} For each task, we show from left to right: \textbf{(a)}~the average number of steps required for a capability to emerge decreases with model scale; \textbf{(b)}~for a single training run, the model's probability for the correct token (highlighted in green) increases sharply in a small interval which we consider the point of emergence; \textbf{(c)}~emergence coincides with learning interpretable attention patterns which we select through our causal ablation; \textbf{(d)}~by manually adjusting the attention map for only $K$ causal heads, we can elicit emergence in the previous checkpoint. Higher $K$ leads to a stronger increase in target token probability, indicating the core attention patterns for each capability are distributed across multiple heads.
      }
    \label{fig:emergence_tasks}
\end{figure}

We extend the string copying experiment illustrated in Figure~\ref{fig:emergence} to three more tasks: in-context repetition, pattern completion, and indirect object identification. Figure~\ref{fig:emergence_tasks} visualizes the results. We provide a complete interpretation of individual task results, along with their relevance to prior work~\citep{olsson2022incontextlearninginductionheads, michaud2024quantizationmodelneuralscaling, wang2022interpretabilitywildcircuitindirect} in Appendix~\ref{app:interp}.

\paragraph{General Trends.}
In all of our emergence experiments (Figures~\ref{fig:emergence}~and~\ref{fig:emergence_tasks}), we observe multiple behaviors consistent with prior work: the emergence of capabilities in a given model depends heavily on the model's initialization~\citep{zhao2026randomscalingemergentcapabilities}, the rate of emergence improves with model scale, and individual capabilities are acquired through discrete phase changes rather than continuously~\citep{michaud2024quantizationmodelneuralscaling}. We additionally demonstrate that capabilities emerge at varying points throughout training depending on the initialization, emergence coincides with learning at least one interpretable task-relevant attention pattern, and larger models acquire capabilities more often and at earlier training steps. Our results indicate that emergent capabilities in language models arise \textbf{randomly and abruptly} over the course of training and depend on \textbf{learning core attention patterns} without which the model could not solve the task. 

Our causal ablations additionally show that attention pattern learning is a \textbf{bottleneck} for language model capabilities, as patching learned attention patterns can elicit emergence in an earlier checkpoint.
Moreover, the causal contribution is not isolated in any single head, but rather distributed across many heads.
These ablations suggest that the improved rate of emergence in larger models may relate to their increased efficiency in learning attention patterns.
We now investigate this hypothesis by extensively studying attention pattern learning in synthetic tasks where the ground-truth pattern is fully specified.

\begin{takeaway}
Language model capabilities emerge abruptly at varying points during training. Emergence coincides with the model learning task-relevant attention patterns, often across several attention heads. Patching these patterns into an earlier checkpoint elicits the capability before it arises naturally, indicating that learning attention patterns is the bottleneck. Larger models overcome this bottleneck earlier and more often across random seeds.
\end{takeaway}

\section{Learning Synthetic Attention-Based Tasks}
\label{sec:synthetic}

Our experiments in Section~\ref{sec:emergence} show that learning attention patterns drives the abrupt acquisition of capabilities in language models, and larger models learn these patterns faster and more reliably. However, there is no single ground-truth attention pattern in natural language tasks, so we cannot directly study the effects of factors such as attention sparsity and context length. We now train transformers on two synthetic attention-based tasks, linear maps and cellular automata, to study these factors in detail.

\begin{figure}[t]
    \centering
    \includegraphics[width=0.8\linewidth]{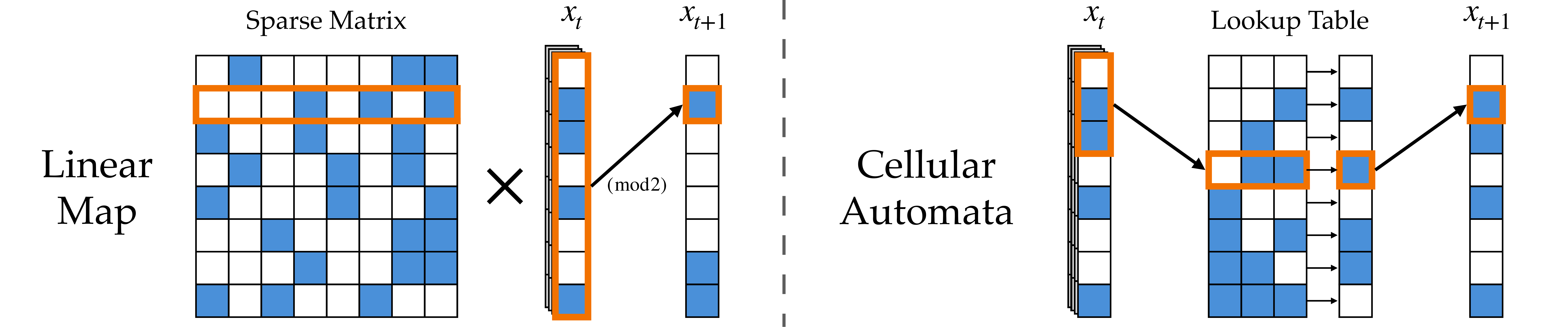}
    \caption{\textbf{Overview of transition dynamics for linear map and cellular automata datasets.} Cells highlighted in orange indicate the relevant information for computing a single cell of the next state. Linear map state transitions compute the parity of a sparse subset of the previous state for each cell. Cellular automata state transitions depend on a deterministic lookup table indexed by a local window of cells. Learning both tasks involves attending to a specific set of cells in the previous state.}
    \label{fig:tasks}
\end{figure}

\subsection{Task Definitions}
We follow a standard language model training setup with a next-token-prediction objective, except instead of text we train models on one of two synthetic tasks.
For both tasks, we define a state size $S$, a trajectory length $T$, and a number of colors $C$. We sample an initial state $x_0=\{0,\dots, C-1\}^S$ and evolve the state $T-1$ times with a recursive function $x_{t+1}=f(x_t)$ to produce a trajectory $[x_0,x_1,\dots, x_{t-1}]$. 
We then train an autoregressive language model with vocabulary size $C$ on the flattened sequence of $ST$ tokens, where the minimum loss is at least $\frac{S-1}{ST}\ln C$ as the model cannot predict the next tokens of the initial random state. 
We provide a table with all the notation used for synthetic tasks in Appendix~\ref{app:notation} for clarity.

\paragraph{Linear Map.} Figure~\ref{fig:tasks} (left) visualizes the transition dynamics for our linear map task. We always use $C=2$ and $T=2$, so that cells contain binary values and the model is trained solely on an initial random state and the subsequent transition state. Formally, the linear map transition function is $f(x)=Ax\bmod 2$ given a sparse matrix $A\in\{0,1\}^{S\times S}$. As shown in Figure~\ref{fig:tasks}, predicting each output bit $x_{t+1, i}$ is equivalent to computing the bitwise parity of a subset of bits in the previous state indexed by $A_{i, :}$. We sample a constant $A$ for each training run, with the constraint that each row of $A$ must contain exactly $s$ nonzero entries for some sparsity level $s\in\{1, 2, \dots, S\}$. Unless otherwise specified, we set $S=16$ and $s=3$ so that the task is sufficiently complex yet still learnable by a transformer (see Figure~\ref{fig:states_linmap}).

\paragraph{Cellular Automata.} Figure~\ref{fig:tasks} (right) demonstrates our process for generating cellular automata datasets. A lookup table $R:\{0,\dots, C-1\}^3\to\{0,\dots, C-1\}$ determines how to compute a single bit of $x_{t+1}$ given a local window $x_{t, i-1:i+1}$. Each lookup table defines a \textit{rule} which maps $r_R(x)_i=R(x_{t, i-1:i+1})$ for all $i\in\{0, 1,\dots, S-1\}$. We introduce two complexity parameters: the number of total rules $N$ and the recursive depth $k$. We sample $N$ lookup tables for each training run, where each lookup table is composed $k$ times for each state transition. To generate a single training example, we sample a single rule and apply the transition function $x_{t+1}=f(x)=r^k(x)$ iteratively starting with a random initial state. When $N=1$, this task involves learning the local window attention pattern and memorizing a static lookup table. However, for $N>1$, the model must first infer the rule $r$ from the prior context $x_{0:t}$ before computing a posterior prediction for $x_{t+1}$.

These two tasks enable us to study two important factors of attention pattern learning in depth. In the linear map task, we vary the \textit{sparsity} of the attention pattern by varying both the state size $S$ and sparsity parameter $s$. We can also precisely measure how accurately a model has learned the ground-truth pattern by comparing to the matrix $A$ that generates the data. The cellular automata task consists of a simple, fixed local window attention pattern, but for $N>1$ rules the model must apply this pattern across multiple states to infer the active rule for each sample. In contrast to the linear map task, we study the effect of \textit{context length} by increasing the state size or trajectory length.

\subsection{Abrupt Learning with Linear Maps}
\label{sec:abrupt}

\begin{figure}[t]
    \centering
    \includegraphics[width=\linewidth]{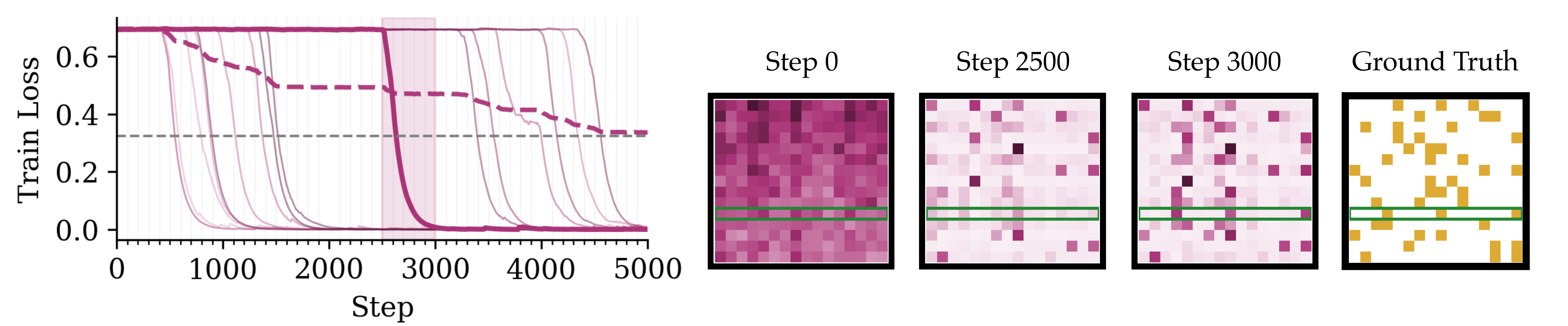}
    \caption{\textbf{Abrupt learning on the linear map task corresponds to learning rows of the ground-truth matrix.} We plot the aggregate loss curve (dashed) decomposed into loss on individual output tokens (light purple) for a 1-layer transformer. We highlight one loss jump in bold and find that it corresponds to a drop in entropy for two attention heads. We plot the averaged attention maps for these two heads at initialization, along with before and after the loss jump (steps 2500 and 3000 respectively). We see that the loss jump corresponds to two attention heads jointly learning a single row of the ground truth linear map.
    }
    \label{fig:jumps}
\end{figure}

We begin by reproducing the abrupt learning phenomenon in Section~\ref{sec:emergence} in our synthetic linear map setting. We train a single layer transformer model with hidden dimension $128$, MLP dimension $512$, $8$ attention heads, and a binary vocabulary on the linear map task with state size $S=16$ and sparsity $s=3$. Solving the sparse linear map directly corresponds to \textbf{learning the ground-truth attention pattern}, as we visualize in Figure~\ref{fig:jumps}. We observe abrupt learning throughout training, characterized by long plateaus in the overall loss followed by sharp jumps. By tracking the loss on each individual output token, we see that loss jumps correspond to abruptly learning a single row of the ground-truth matrix $A$. 

We also track the entropy $-\sum_{i=S}^{2S-1}\sum_{j=0}^{S-1} s_{ij}\log s_{ij}$ for each head across queries $x_{S:2S-1}$ and keys $x_{0:S-1}$, where $s_{ij}$ is the batch-averaged softmax score for query position $i$ and key position $j$. In Figure~\ref{fig:jumps}, the highlighted loss jump corresponds to a sharp decrease in entropy for two attention heads~(see Appendix~\ref{app:entropy}). We visualize the attention heads before and after the loss jump and observe a transition from roughly uniform attention toward the ground-truth pattern for a single row. We note this abrupt learning is also \textit{random}; different initialization seeds learn the sparse linear map rows in different orders and at different times.

Since we define the ground-truth attention pattern by the matrix $A$, we can apply the attention intervention proposed by~\citet{gopalani2025happenslossplateauunderstanding}. We apply an attention bias to each attention head, where we add $c\cdot A$ to the attention logits for queries $x_{S:2S-1}$ and keys $x_{0:S-1}$ before computing softmax scores. We find that this intervention enables the model to learn the linear map task smoothly and almost instantly, with loss curves resembling those of a single jump in Figure~\ref{fig:jumps}. We provide the complete training results in Appendix~\ref{app:intervention}. This immediate learning under the attention intervention indicates that the linear map task is \textbf{attention-heavy}: the main bottleneck is in learning the correct attention pattern, and the subsequent computation of the dot-product $\bmod\ 2$ is relatively trivial.

\begin{figure}[t]
  \centering
  \begin{subfigure}[b]{0.3\textwidth}
    \centering
    \includegraphics[width=\textwidth]{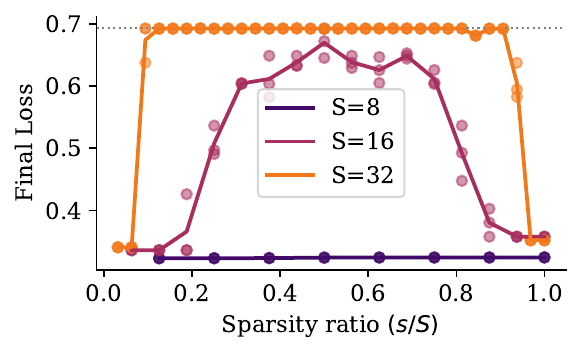}
    \caption{Linear Map}
    \label{fig:states_linmap}
  \end{subfigure}
  \hfill
  \begin{subfigure}[b]{0.3\textwidth}
    \centering
    \includegraphics[width=\textwidth]{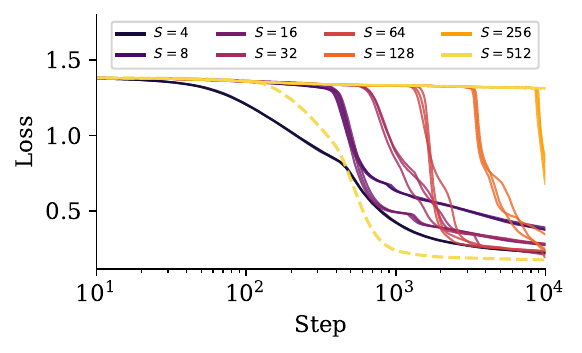}
    \caption{Cellular Automata} 
    \label{fig:states_ca}
  \end{subfigure}
  \hfill
  \begin{subfigure}[b]{0.25\textwidth}
    \centering
    \includegraphics[width=\textwidth]{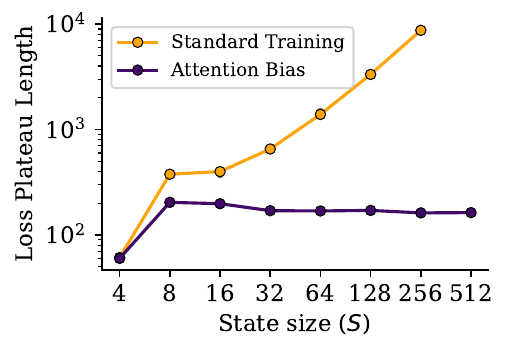}
    \caption{Loss Plateau} 
    \label{fig:plateau}
  \end{subfigure}
  \caption{\textbf{Training dynamics become abrupt with longer context and medium sparsity.} We average training loss over three models trained with different seeds. \textbf{(a)} Models can always learn very sparse or very dense attention patterns in the linear map task, but as we increase the state size medium sparsity patterns become more difficult and unlearnable by $S=32$. \textbf{(b)} Given a fixed token batch size, increasing the state size $S$ and decreasing the number of samples per batch shifts the loss dynamics from smooth to abrupt. Dashed line: biasing the model's attention scores with the ground-truth pattern results in immediate learning for $S=512$; without this intervention, the model makes no progress after 10,000 steps. \textbf{(c)} Loss plateau length increases multiplicatively with state size, but applying our attention intervention neutralizes this effect.}
  
  \label{fig:state_size}
\end{figure}

We now study how context length and attention sparsity affect the difficulty of learning the linear map task. We vary the state size $S$ to control context length, and sweep the sparsity ratio $s/S$, which controls what proportion of the input state the model must attend to for predicting each output bit. In Figure~\ref{fig:states_linmap} we illustrate final loss as a function of the ratio $s/S$, sweeping across all $s\in\{1,2,\dots, S\}$ for $S\in\{8, 16, 32\}$. We see that for a small state size ($S=8$), the model can learn any linear map regardless of the sparsity level. However, at $S=16$ the task becomes more difficult, especially for linear maps with medium sparsity $s/S\approx0.5$. At $S=32$, the task is almost always unlearnable after 10,000 steps except for the simplest attention patterns ($s=1$, $s=S$). We generally see that the difficulty of learning an attention pattern depends jointly on sparsity and context length; even extremely sparse or dense patterns become unlearnable at larger $S$. We now demonstrate the same effect in the cellular automata task, where the attention pattern is simpler but increasing the context length has a multiplicative effect on learning difficulty.

\subsection{Long-Context Attention Patterns are Difficult to Learn}
\label{sec:context}

In the cellular automata task, the model must attend to a fixed local window of the previous state in order to predict each subsequent output cell~(Figure~\ref{fig:tasks}). However, even this relatively simple pattern is made arbitrarily difficult by increasing the state size. In Figure~\ref{fig:states_ca} we sweep state sizes $S\in\{4, 8,16,\dots,256\}$, using $C=4$ colors, $T=16$ trajectory length, $N=256$ rules, and recursive depth $k=1$. 
We train a 4-layer model for cellular automata, as a single layer is not sufficient to solve the multi-rule task. The task is relatively easy for small $S$, but as we increase $S$ the loss curves become more abrupt. As the state size grows, identifying the relevant local window within a longer sequence becomes increasingly difficult, even though the pattern itself remains simple. With $S=512$, no model makes progress after 10,000 steps. Although the underlying algorithm generating the data is identical, varying the context length determines whether a model can easily learn the task or fail to make any progress at all.

We again apply an attention intervention where we artificially inflate attention logits in each head based on a ground-truth pattern. For this task, we encourage the model to attend to the window of cells from the previous state for each token, as illustrated in Figure~\ref{fig:tasks}. In Figure~\ref{fig:plateau}, we measure the loss plateau length as the step at which the model first achieves loss $<1.3$. We see that increasing the state size has a multiplicative effect on the length of the loss plateau, but with our attention intervention the plateau length is negligible ($\approx 100$ steps) and remains constant regardless of context length. Our intervention confirms that \textbf{learning a long-context attention pattern} is a bottleneck to training on our cellular automata task.

In Appendix~\ref{app:plateau} we also vary the context length by adjusting the trajectory length given a fixed state size. Similar to Figure~\ref{fig:plateau}, increasing trajectory length leads to longer loss plateaus, although the relationship is non-monotonic as the model struggles to learn the cellular automata dynamics when the trajectory is too short. We also vary the number of rules $N$, the recursive depth $k$, and cell colors $C$; notably, these variables do not affect the loss plateau length, suggesting that context length is the primary factor.

\begin{takeaway}
On synthetic tasks where the ground-truth attention pattern is known, the difficulty of learning that pattern is governed by context length and sparsity. The same task becomes unlearnable as we lengthen the context, and medium-sparsity patterns are the hardest to find. Biasing attention scores toward the ground-truth pattern removes the loss plateau entirely, confirming that learning the token-mixing pattern is the bottleneck for training.
\end{takeaway}

\section{The Role of Architecture in Learning Attention Patterns}
\label{sec:arch}
We now investigate how model architecture interacts with learning attention patterns in our synthetic tasks. In all experiments we use a baseline with $D=128$ with $H=8$ attention heads. We use a single-layer model for the linear map task and a $4$-layer model for cellular automata. Surprisingly, deeper models perform worse on the linear map task, likely because composing attention layers makes it harder to express a single sparse pattern directly; we provide additional details in Appendix~\ref{app:depth}. In Sections~\ref{sec:width}~and~\ref{sec:heads} we study the effect of scaling width and show that including more attention heads improves performance on both synthetic tasks. Finally, in Section~\ref{sec:alt_arch} we see that architectures with alternative attention mechanisms significantly outperform transformers on the linear map task, but this benefit does not translate to cellular automata.

\begin{figure}[t]
    \centering
    \includegraphics[width=1\linewidth]{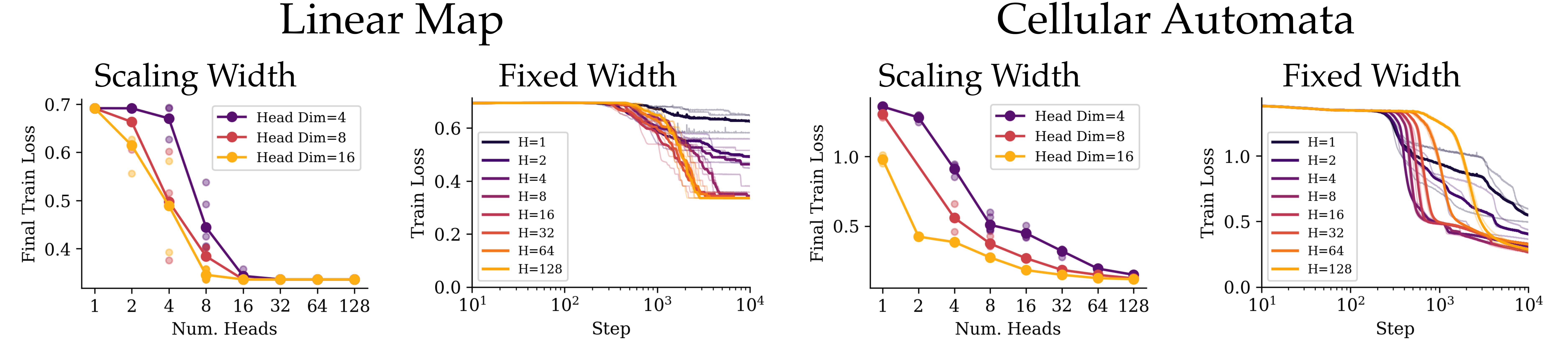}
    \caption{\textbf{Scaling heads always improves performance on the linear map task, but cellular automata benefits from a minimum head dimension.} For both tasks, we scale the width (left) by increasing the number of heads $H$ with three head dimensions (4, 8, and 16). Given a fixed width (right), we also sweep over $H$, restricting the head dim to $128/H$. A transformer with head dimension 1 and $H=128$ heads quickly learns the linear map task; however, it is much slower in learning the cellular automata task compared to a model with fewer heads and a higher head dimension.}
    \label{fig:width}
\end{figure}

\subsection{Scaling Width}
\label{sec:width}
We scale the width for single-layer models on the linear map task and $4$-layer models on the cellular automata task. In Figure~\ref{fig:width} we vary the width by scaling the number of heads $H$ with head dimensions $4$, $8$, and $16$. For both synthetic tasks, \textbf{increasing the number of heads consistently reduces the final loss}. We hypothesize that each attention head independently searches over possible patterns, so more heads increase the probability of finding the correct pattern. For cellular automata, we also observe that increasing the head dimension reduces training loss for a fixed number of heads $H$; we believe this is because each head needs sufficient capacity to store rule information from prior context. This tradeoff raises the question of how to optimally allocate capacity between head count and head dimension at a fixed model width.

\subsection{Scaling Attention Heads}
\label{sec:heads}
To control for the number of attention heads in isolation, we fix the width to $D=128$ and sweep $H=1,2,4,\dots, 128$ heads with head dimension $128/H$. In Figure~\ref{fig:width} we see that a model with $H=128$ heads and head dimension $1$ can still solve the linear map task, confirming that head count matters more than head capacity for attention-heavy tasks. 
However, in the cellular automata task, setting $H=128$ results in significantly slower learning and slightly higher final loss, likely due to the reduced capacity for storing rule information. 
For both tasks, the optimal configuration favors many heads with a small head dimension, but there is a task-dependent minimum head dimension below which increasing head count is no longer beneficial.

\subsection{Alternative Architectures}
\label{sec:alt_arch}
\begin{figure}[t]
  \centering
  \hspace*{0.1\textwidth}
  \begin{subfigure}[b]{0.3\textwidth}
    \centering
    \includegraphics[width=\textwidth]{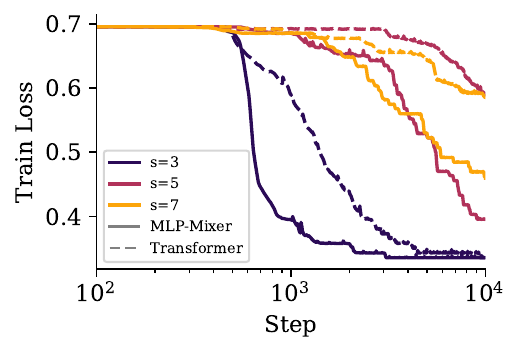}
    \caption{Linear Map}
    \label{fig:arch_linmap}
  \end{subfigure}
  \hfill
  \begin{subfigure}[b]{0.3\textwidth}
    \centering
    \includegraphics[width=\textwidth]{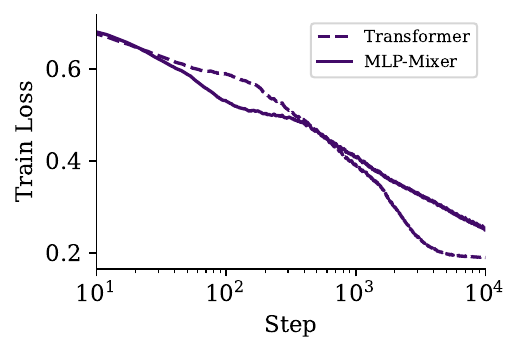}
    \caption{Cellular Automata} 
    \label{fig:arch_ca}
  \end{subfigure}
  \hspace*{0.1\textwidth}
    \caption{\textbf{MLP-Mixer outperforms a transformer on the sparse linear map task, while underperforming on cellular automata.} As we increase the sparsity for the linear map task with state size $S=16$, the transformer model struggles to learn the task while the MLP-Mixer achieves much lower loss. Note $s=7$ corresponds to a maximally difficult attention pattern, as illustrated in Figure~\ref{fig:states_linmap}.
    }
    \label{fig:arch}
\end{figure}

We now compare a standard transformer to MLP-Mixer~\citep{tolstikhin2021mlpmixerallmlparchitecturevision}. MLP-Mixer replaces dot-product attention with a static learned matrix that mixes information across sequence positions. In Figure~\ref{fig:arch}, we show that \textbf{MLP Mixer learns the linear map task significantly faster than a transformer}, demonstrating that it is possible to design architectures that outperform transformers on attention-heavy tasks. However, MLP Mixer underperforms on the cellular automata task, likely because it must learn the local-window pattern independently at each position, whereas dot-product attention can parameterize it directly. This performance gap motivates designing architectures that learn positional patterns as efficiently as MLP-Mixer while retaining a transformer's ability to learn context-dependent token-mixing patterns.

We also compare the transformer to architectures with alternative token-mixing patterns, including Mamba~\citep{gu2024mambalineartimesequencemodeling}, Gated DeltaNet~\citep{yang2025gateddeltanetworksimproving}, and RWKV~\citep{peng2023rwkvreinventingrnnstransformer}. All other architectures outside of MLP-Mixer underperform relative to transformers on both the linear map and cellular automata tasks; full details are in Appendix~\ref{app:arch}.

\begin{takeaway}
Architectural choices significantly impact how efficiently a model learns token-mixing patterns. Scaling the number of attention heads speeds up pattern learning across both synthetic tasks. Alternative token-mixing mechanisms can yield further gains, as MLP-Mixer learns the linear map task an order of magnitude faster than a transformer.
\end{takeaway}

\section{Related Work}

\paragraph{Emergent Capabilities.}
Many downstream capabilities of language models emerge sharply past certain model scales~\citep{wei2022emergentabilitieslargelanguage}. \citet{schaeffer2023emergentabilitieslargelanguage} argue emergence is a mirage caused by discontinuous metrics. We show that, even on the continuous metric of correct-token probability, capabilities arise sharply throughout training and across model scales. \citet{zhao2026randomscalingemergentcapabilities} show emergence is random across initialization seeds at a fixed scale; we confirm these results on Pythia models and further demonstrate that emergence is random across training steps. \citet{zucchet2025emergencesparseattentionimpact} study how sparse attention patterns relate to emergence, developing a theoretical framework for how task structure governs emergence in small transformer models. We complement this framework with causal evidence in pretrained language models, showing that patching a small set of attention heads can directly elicit capabilities before they arise naturally. We further isolate the factors that make a pattern hard to learn, namely context length and sparsity, and show that architectural choices control the rate at which these patterns are learned.

\paragraph{Abrupt Learning.}
Training transformers on algorithmic tasks is known to produce abrupt learning, where loss plateaus end in sudden drops~\citep{gopalani2024abruptlearningtransformerscase}. \citet{gopalani2025happenslossplateauunderstanding} show that loss plateaus persist until the model learns the correct attention map, at which point training converges rapidly; intervening on attention maps can trigger this transition earlier. Abrupt learning also appears in natural language pretraining: \citet{chen2025suddendropslosssyntax} observe abrupt learning in BERT when specific attention heads learn syntactic structure, and \citet{olsson2022incontextlearninginductionheads} show that induction heads in autoregressive models produce a sharp phase transition in in-context learning ability. \citet{kangaslahti2026hiddenbreakthroughslanguagemodel} develop an unsupervised method for detecting these hidden breakthroughs from smooth pretraining loss curves. We note abrupt learning is distinct from grokking~\citep{power2022grokkinggeneralizationoverfittingsmall, nanda2023progressmeasuresgrokkingmechanistic}; abrupt learning is a sudden drop in training loss, while grokking is a sudden improvement in generalization after training loss has already converged.

\paragraph{Neural Scaling Laws.} 
While language model capabilities emerge randomly and abruptly, pretraining loss decreases smoothly as a power law with respect to model scale and dataset size~\citep{hestness2017deeplearningscalingpredictable,kaplan2020scalinglawsneurallanguage}. The quantization model of neural scaling~\citep{michaud2024quantizationmodelneuralscaling} reconciles these two observations, hypothesizing that language models learn discrete skills (quanta) abruptly, and that the aggregate of many such phase changes produces smooth scaling curves. Our findings support this view; we further demonstrate that phase changes correspond to the abrupt learning of attention patterns, and improved emergence rates in larger models are due to acquiring these quanta faster on average.

\paragraph{Our Contributions.}
Prior work establishes that emergence is abrupt~\citep{wei2022emergentabilitieslargelanguage}, random across initializations~\citep{zhao2026randomscalingemergentcapabilities}, and connected to attention~\citep{olsson2022incontextlearninginductionheads, chen2025suddendropslosssyntax}. We unify and extend these observations in four directions. First, we show emergence is random across training \textit{steps}, not only across initialization seeds, and we pinpoint the step at which it occurs for multiple models. Second, we provide causal evidence showing that patching a handful of attention heads elicits a capability before it arises naturally. Third, we design synthetic tasks with known ground-truth patterns to isolate context length and sparsity as the factors that govern how hard a pattern is to learn. Finally, we show that architectural choices such as attention head count and the token-mixing mechanism directly control attention-pattern learning, with MLP-Mixer learning our linear map task an order of magnitude faster than a transformer.

\section{Conclusion}

Emergent capabilities in language models arise randomly and abruptly because learning sparse attention patterns is inherently difficult for transformers. Our causal ablations on Pythia models show that patching a small set of attention heads can elicit a capability before it naturally emerges, demonstrating that attention pattern learning is a bottleneck for emergence. Our synthetic training experiments reveal that the difficulty of learning an attention pattern depends on context length and pattern sparsity. Scaling head count accelerates attention pattern learning, likely because more heads provide more candidates for discovering the correct pattern. We show alternative architectures like MLP-Mixer outperform transformers specifically when the task requires learning a complex positional attention pattern. Our findings prescribe multiple potential future directions for long-context training: attention interventions during training, attention mechanisms with an inductive bias towards sparsity, and distillation of attention maps from stronger models.

Our Pythia experiments in Section~\ref{sec:emergence} are limited to models below 1B parameters, and the tasks we study involve relatively elementary capabilities. Whether our findings generalize to more sophisticated capabilities and larger models is an open question. Additionally, our synthetic tasks involve positional attention patterns; context-dependent patterns such as induction heads may exhibit qualitatively different properties. We also believe synthetic pre-pretraining~\citep{hu2025circuitschomskyprepretrainingformal, lee2026traininglanguagemodelsneural} may accelerate the attention pattern search by seeding the model with useful inductive biases before natural language training. We also provide preliminary results for testing whether pre-pretraining affects the acquisition of emergent capabilities in Appendix~\ref{app:ppt}. We hope these results motivate future work on designing datasets, architectures, and training procedures that make attention pattern learning faster and more reliable.

\paragraph{Acknowledgements.} We thank Lily Li and Martin Marek for helpful discussions.
We also thank NSF CAREER IIS-2145492, NSF CDS\&E-MSS 2134216, and DARPA AIQ HR00112590066 for support.

\bibliography{references}
\bibliographystyle{plainnat}

\newpage
\appendix

\section{Emergent Capabilities in Pythia Language Models}
\label{app:causal}

\subsection{Causal Ablations and Attention Sinks}
We ablate an individual attention head as follows. In the pre-emergence model, the output for an attention head at a given head and layer is $SV$, where $S\in\mathbb{R}^{N\times N}$ is the attention score matrix computed by softmax over key-value dot products, and $V\in\mathbb{R}^{N\times D}$ is the set of value vectors for the given head. We instead compute $S'V$, where $S'$ is the attention score matrix for the given head resulting from passing the prefix $x$ to the post-emergence model. We then compute the forward pass through the rest of the layers in the pre-emergence model, and measure the change in the output token probability for the ground-truth token. We select causal heads by applying this ablation to each individual head, and sorting by the change in correct token probability.

When applying our causal ablation, especially in early checkpoints, many top causal heads exhibit attention sink behavior~\citep{xiao2024efficientstreaminglanguagemodels} where every query attends to the first token. We filter these out from our analysis before applying the top-$K$ filtering to focus on heads that display interpretable behavior. One way to define attention heads is by testing whether every query token attends to key position 0 with the highest weight. However, we find that this is not sufficient to filter out sink-like behavior as attention sinks gradually emerge over time; this criteria does not include heads which are in the process of becoming attention sinks but do not yet place maximal attention weight on token 0. We instead define an attention sink in the context of our ablation as a head where the attention weight to key position zero \textit{increases for each query token} between the pre-emergence and post-emergence checkpoints. If the attention weight to position zero for an attention head increases for \textit{all} queries in our sample, we discard this head from our analysis. This allows us to find attention pattern transitions where a head learns to attend to some token position other than 0 for \textit{any} query.

\subsection{Analyzing Emergence on Multiple Samples}
\label{app:sample}

In Section~\ref{sec:emergence} we show that the model's probability of predicting the correct next token for a single sample in a given task spikes abruptly throughout training. Although this is the case for individual samples, we find that emergence times vary for different samples within the same task, which makes it difficult to probe for causal attention heads and measure trends across model scale. We speculate that this may be due to distinct circuits in the model that decide when to apply an algorithm on a given sample and the specific algorithm to apply; the former may be learned at different rates even if the latter is learned at once. To test this hypothesis, we run our emergence analysis on multiple samples of the copying task, and observe how many causal attention heads are consistently shared across tasks. Specifically, we apply our forward causal head ablation on individual samples, track the $K=16$ top heads for each sample, and measure the degree to which the top $K=16$ heads overlap across pairs of samples. 

We measure overlap using the intersection over union (IOU) of causal head sets; Table~\ref{tab:iou} shows the results. We see that there is a high degree of overlap for causal attention heads in the repetition task for most models --- much higher than what we would expect if the model were utilizing random heads for each sample. Note we only include seeds for which the model exhibits the copying capability across all samples. 

\begin{table}[h]
\centering
\caption{\textbf{Overlap between causal heads for different samples of the copying task for Pythia 410m.} We select the top $K=16$ causal heads across multiple samples and compare the set overlaps between pairs of samples. Most models utilize many of the same attention heads to solve the copying task, even when the correct token probability spikes at different training steps for each sample. We compare to a random baseline $K / (2N-K)$, where $N$ is the number of non-sink heads from which we apply our causal ranking.}
\label{tab:iou}
\resizebox{\textwidth}{!}{%
\rowcolors{2}{gray!10}{white}
\begin{tabular}{lccccccc|c}
\toprule
Metric & Default & Seed $1$ & Seed $4$ & Seed $5$ & Seed $6$ & Seed $7$ & Seed $9$ & Average \\
\midrule
Total Heads & $22$ & $27$ & $24$ & $25$ & $28$ & $28$ & $30$ & $26$ \\
Common Heads & $10$ & $5$ & $8$ & $7$ & $4$ & $4$ & $2$ & $6$ \\
IOU & $0.455$ & $0.185$ & $0.333$ & $0.280$ & $0.143$ & $0.143$ & $0.067$ & $0.229$ \\
Random Baseline & $0.060$ & $0.031$ & $0.054$ & $0.073$ & $0.039$ & $0.034$ & $0.048$ & $0.048$ \\
Ratio & $7.614$ & $6.007$ & $6.229$ & $3.815$ & $3.625$ & $4.241$ & $1.383$ & $4.740$ \\
\bottomrule
\end{tabular}%
}
\end{table}

\subsection{Interpreting Causal Attention Mechanisms}
\label{app:interp}
In Figures~\ref{fig:emergence}~and~\ref{fig:emergence_tasks} we apply our emergence analysis across four tasks: copying, in-context repetition, pattern completion, and indirect object identification. Below, we interpret the qualitative findings in each figure and how they relate to established prior results:

\paragraph{Copying.}
In the copying task we provide the model with a string followed by all but the last token of the same string as context, and we evaluate whether the model can predict the last token of the repeated string. For models that learn to solve the copying task, the capability emerges sharply, and two of the causal attention heads display patterns corresponding to copying and attending to the previous token.

\paragraph{In-Context Repetition.} 
\citet{olsson2022incontextlearninginductionheads} empirically demonstrate the presence of induction heads in language models, which implement a primitive algorithm used for in-context learning. Induction heads are responsible for completing the sequence \texttt{[A] [B] \dots [A] $\rightarrow$ [B]}, by searching for previous instances of phrase \texttt{[A]} and predicting that \texttt{[B]} will follow after. In our example, \texttt{[A]}  and \texttt{[B]} are the strings ``atten-'' and ``tion'' respectively. We see this repetition capability emerges rather early during training even for smaller models, and performance on this task follows a phase change-like transition as shown by~\citet{olsson2022incontextlearninginductionheads}.

\paragraph{Pattern Completion.}
The quantization model of neural scaling~\citep{michaud2024quantizationmodelneuralscaling} posits that the smooth aggregate performance of neural networks can be decomposed into discrete skills (quanta) that are learned abruptly, and the authors find that one empirical quanta in Pythia models corresponds to predicting next entries of numbered lists. We confirm this skill is learned abruptly in Figure~\ref{fig:emergence_tasks}. We additionally observe that larger models acquire this skill faster even when trained on the same data distribution and the discrete phase change is a result of the abrupt learning of attention patterns.

\paragraph{Indirect Object Identification.}
We probe for indirect object identification using the example illustrated by~\citet{wang2022interpretabilitywildcircuitindirect}. We visualize one selected attention head in Figure~\ref{fig:emergence_tasks}, which roughly corresponds to the name-mover behavior demonstrated by~\citet{wang2022interpretabilitywildcircuitindirect} in GPT-2-small. Even at the scale of GPT-2-small (124M), this capability is often not present in most model seeds. Note that the visualized head demonstrates the \textbf{correction} of an attention pattern; before emergence, the last query token places significant weight on ``John'', but after emergence most of the weight collapses onto the correct token ``Mary''.   

\section{Abrupt Learning on Synthetic Tasks}
\subsection{Notation}
\label{app:notation}
In Table~\ref{tab:notation} we clarify notation for all of the symbols we refer to when describing our synthetic tasks.
\begin{table}[b]
\centering
\rowcolors{2}{gray!10}{white}
\resizebox{\linewidth}{!}{
\begin{tabular}{ll}
\toprule
\textbf{Symbol} & \textbf{Description} \\
\midrule

\rowcolor{gray!25}
\multicolumn{2}{l}{\textbf{General notation}} \\
$C$ & Number of colors; number of possible values each cell can take \\
$S$ & State size; number of cells in a single state \\
$T$ & Trajectory length; total number of states in a sequence \\
$f$ & Transition function; maps a state to the next state in the trajectory \\

\midrule
\rowcolor{gray!25}
\multicolumn{2}{l}{\textbf{Linear map setting ($C=2$, $T=2$)}} \\
$A \in \{0,1\}^{S \times S}$ & Sparse discrete linear map; each entry of $Ax$ is a bitwise parity over selected entries of $x$ \\
$s \in \{1,\dots,S\}$ & Sparsity of $A$; number of nonzero entries per row \\

\midrule
\rowcolor{gray!25}
\multicolumn{2}{l}{\textbf{Cellular automata}} \\
$R : \{0,\dots,C-1\}^3 \to \{0,\dots,C-1\}$ & Local lookup table mapping a neighborhood to a single cell value \\
$r_R(x)$ & Rule induced by $R$; applies $R(x_{i-1:i+1})$ across all positions to produce the next state \\
$r_R^k(x)$ & $k$-step evolution under rule $R$ (recursive application) \\
$N$ & Number of rules; one rule is sampled per training example \\

\bottomrule
\end{tabular}
}
\caption{Summary of notation used throughout the paper for our synthetic tasks.}
\label{tab:notation}
\end{table}

\subsection{Attention Entropy}
\label{app:entropy}

\begin{figure}[htbp]
  \centering
  \begin{subfigure}[b]{0.45\textwidth}
    \centering
    \includegraphics[width=\textwidth]{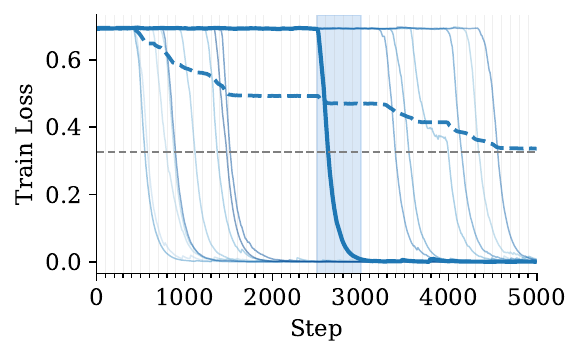}
    \caption{Loss}
  \end{subfigure}
  \hfill
  \begin{subfigure}[b]{0.45\textwidth}
    \centering
    \includegraphics[width=\textwidth]{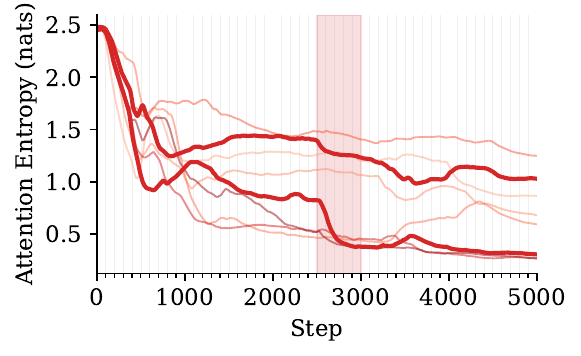}
    \caption{Entropy} 
  \end{subfigure}
  \caption{\textbf{Loss jumps in the linear map task correspond to drops in attention entropy.} In a short window of training, we see \textbf{(a)} the loss for a single output position drops to zero abruptly, and \textbf{(b)} the entropy of two attention heads drops at the same time. Upon inspecting these attention heads in Figure~\ref{fig:jumps} we see that these two attention heads together learn a single row of the ground-truth linear map.}
  \label{fig:entropy}
\end{figure}

In Figure~\ref{fig:entropy} we plot the loss and attention entropy across all eight heads in a single-layer transformer trained on the linear map task. We highlight a window of training steps in which the loss for a single row drops to zero. Within that same window, we see the attention entropy of two heads drop in tandem. As we show in Section~\ref{sec:abrupt}, these two heads learn interpretable attention patterns corresponding to a single row of the ground truth linear map.

\subsection{Attention Interventions}
\label{app:intervention}

\begin{figure}[htbp]
  \centering
  \begin{subfigure}[b]{0.45\textwidth}
    \centering
    \includegraphics[width=\textwidth]{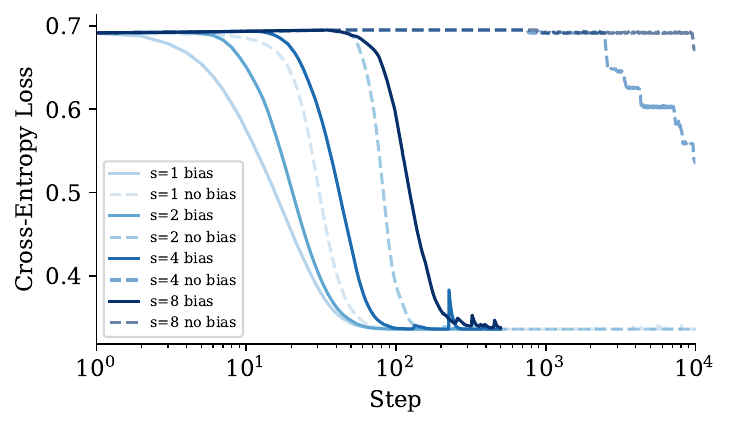}
    \caption{Linear Map}
    \label{fig:bias_linmap}
  \end{subfigure}
  \hfill
  \begin{subfigure}[b]{0.45\textwidth}
    \centering
    \includegraphics[width=\textwidth]{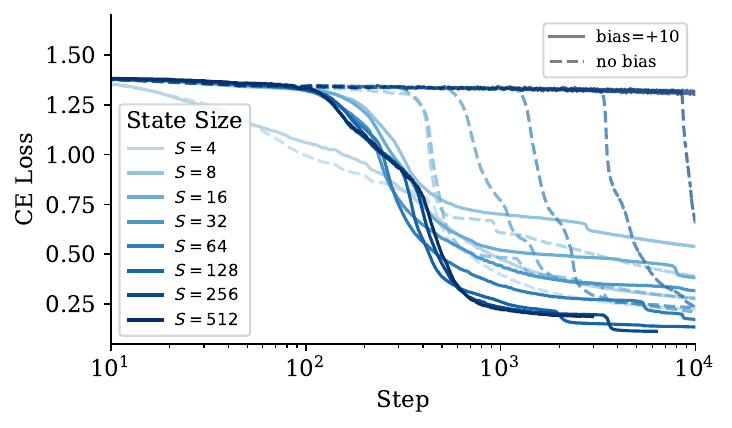}
    \caption{Cellular Automata} 
    \label{fig:bias_ca}
  \end{subfigure}
  \caption{\textbf{Attention intervention enables near-immediate learning on the linear map and cellular automata tasks.} We bias the attention scores for every attention head based on a ground-truth pattern corresponding to each task, and find that in every case the subsequent learning dynamics are no longer abrupt. }
  \label{fig:bias}
\end{figure}

We apply an attention intervention similar to~\citet{gopalani2025happenslossplateauunderstanding} by first defining a ground-truth attention pattern~(see Figure~\ref{fig:tasks} and Section~\ref{sec:synthetic}) and artificially increasing a model's attention scores based on this pattern. Figure~\ref{fig:bias} illustrates the results. Notably, with $s=8$ and $S=16$ a model makes almost no progress after 10,000 steps, while the intervention enables training convergence in under 1,000 steps. The same is true for cellular automata with $S=512$. 

\subsection{What Variables Affect Loss Plateau Length?}
\label{app:plateau}

\begin{figure}[t]
    \centering
    \includegraphics[width=1\linewidth]{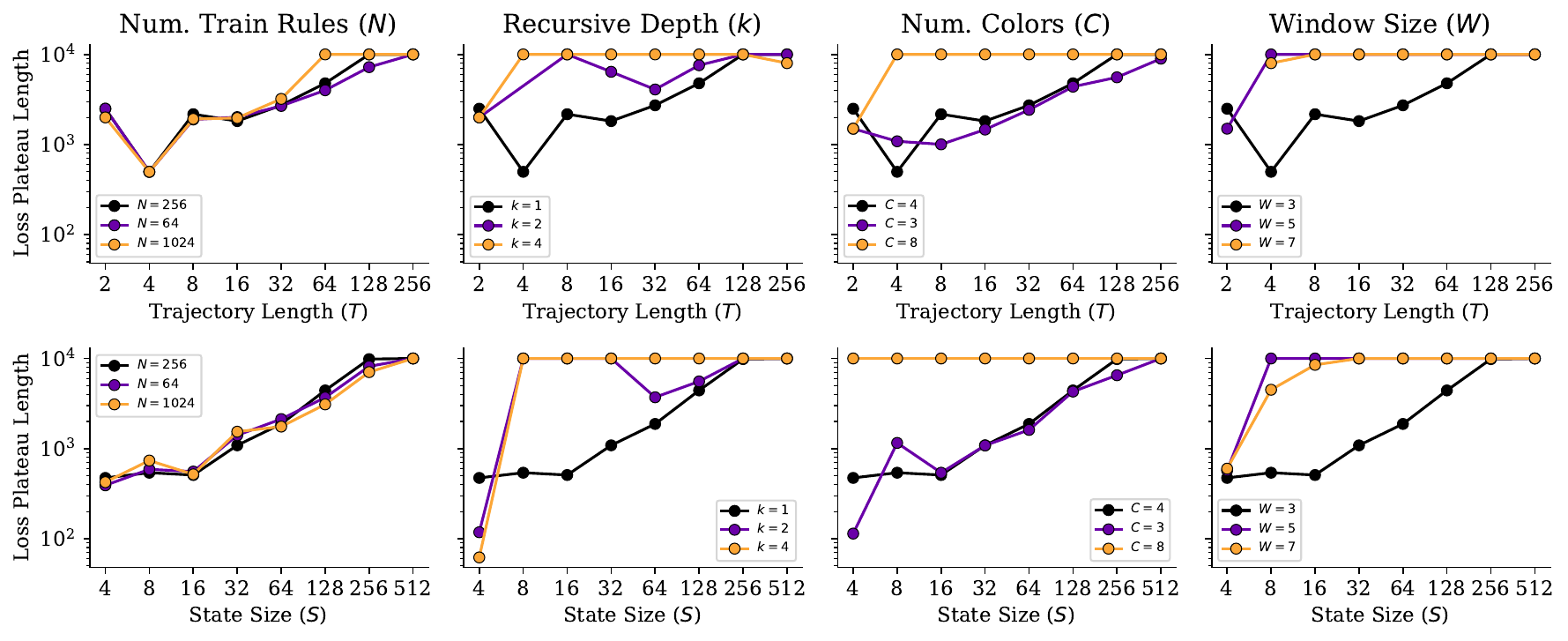}
    \caption{\textbf{Context length is the dominating factor in loss plateau length.} We adjust cellular automata parameters $S, T, N, k, C, W$ and track the length of the initial loss plateau. In each grid cell, we either fix the state size to $S=64$ and vary trajectory length $T$, or fix $T=16$ and vary $S$. In all cases, we adjust the sample batch size $B$ so the total number of tokens per step $S\cdot T\cdot B$ remains fixed. Black lines indicate the baseline config ($N=256$ rules, $k=1$ recursive depth, $C=4$ colors, $W=3$ window size) and are identical across columns in each row. Varying $N, k, C, W$ does not affect the loss plateau length, unless the task becomes too difficult, in which case the plateau length saturates. However, increasing $S$ and $T$ extends the loss plateau, indicating that longer context constitutes a bottleneck to learning the cellular automata task.}
    \label{fig:gca_plateau}
\end{figure}

In Section~\ref{sec:context} we showed that long context lengths exacerbate loss plateaus during training, although we only ablated the effect of state size in Figure~\ref{fig:plateau}. In this section, we also extend our context length analysis by varying the trajectory length $T$, along with other cellular automata parameters which are orthogonal to the context length: number of rules $N$, recursive depth $k$, number of colors $C$, and window size $W$. Note we expect $N, k, C, W$ to still affect the overall difficulty of the task (for example, the amount of training required to achieve the minimum loss), but here we only study the initial loss plateau --- the amount of training steps required to make any progress at all. Figure~\ref{fig:gca_plateau} illustrates our full sweep of the loss plateau length across various cellular automata configurations.

Unlike the relationship to state size shown in Figure~\ref{fig:plateau}, loss plateau length does not always increase monotonically with trajectory length $T$. With a state size of $S=64$ and trajectory length $T=2$, for example, the model only ever sees $S\cdot (T-1)=64$ cell transitions in a single sample, which is not nearly sufficient to exactly infer which of the $N$ rules is active from the context. Thus we see an initial drop in the loss plateau length in the top row of Figure~\ref{fig:gca_plateau}, as increasing $T$ initially provides more useful information to the model. However, increasing $T$ further makes the task more difficult through increasing the context length, hence the longer loss plateaus for large $T$.

We see that in most cases, varying cellular automata parameters outside of $S$ and $T$ does not affect the loss plateau length. Any differences from the baseline trend (black lines) in Figure~\ref{fig:gca_plateau} are due to the loss plateau length saturating at $10,000$ steps (the entire duration of training) as the task becomes too difficult to train on. For example, increasing the number of colors to $C=8$ leads to $C^{C^W}=8^{8^3}$ possible rules, which makes the task too difficult except for the most extreme case of $T=2$. Outside of these unlearnable configurations, we see that \textbf{increasing context length is the primary fa}

\subsection{Depth Scaling}
\label{app:depth}

\begin{figure}[t]
  \centering
  \begin{subfigure}[b]{0.45\textwidth}
    \centering
    \includegraphics[width=\textwidth]{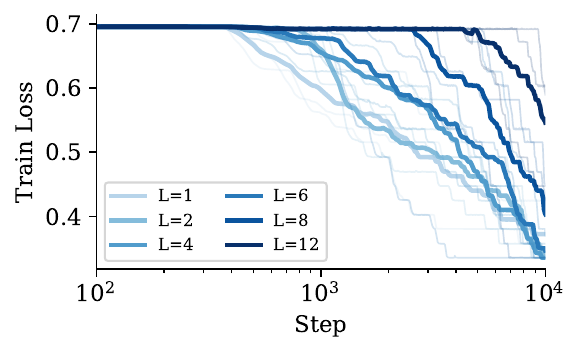}
    \caption{Linear Map}
    \label{fig:depth_linmap}
  \end{subfigure}
  \hfill
  \begin{subfigure}[b]{0.45\textwidth}
    \centering
    \includegraphics[width=\textwidth]{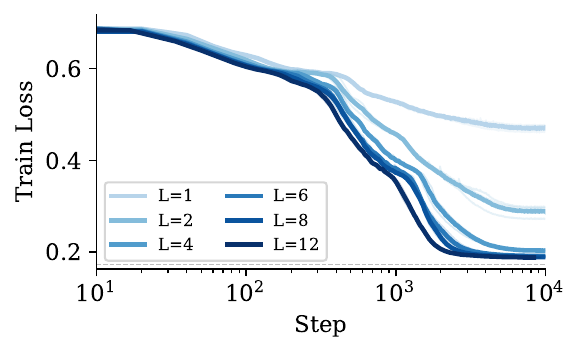}
    \caption{Cellular Automata} 
    \label{fig:depth_ca}
  \end{subfigure}
  \caption{\textbf{Increasing number of layers hurts linear map performance but improves cellular automata performance.} Bold lines indicate average loss over three random seeds. A 1-layer model reliably solves the linear map task, but increasing the number of layers either has no effect or dramatically worsens performance. An 8 or 12 layer model cannot solve the task after 10,000 training steps. In contrast, a single layer is insufficient for the cellular automata task, and scaling the number of layers yields a monotonic improvement in final loss and convergence speed.}
  \label{fig:depth}
\end{figure}

We study the effect of scaling depth when we fix the hidden dimension to $D=128$ and the number of attention heads to $H=8$. In Figure~\ref{fig:depth} (left) we show that \textbf{deeper models perform worse} than shallow models on the linear map task. In fact, there is no significant benefit to scaling beyond a single layer model. We believe this is due to the attention-heavy nature of the linear map task; as visualized in Figure~\ref{fig:jumps}, the model must express an intricate sparse pattern in its attention layers. Composing attention layers recursively may increase the difficulty of learning the ground truth pattern, despite the overall increase in model parameters.

In our cellular automata task, increasing layers has no benefit in the single-rule setting ($N=1$), even if we increase the recursive depth $k$. However, in the multi-rule setting with $N=64$, we see a clear improvement with more layers. Additionally, increasing $k$ with $N=64$ widens the gap between deeper and shallower models. In Figure~\ref{fig:depth} (right) we visualize the effect of scaling layers with $N=64$ rules and depth $k=4$. The multi-rule cellular automata task inherently requires recursive computation, in contrast to the linear map task which does not benefit from increased depth. The cellular automata task resembles a more practical setting for scaling depth, as a task like language modeling requires both learning key attention patterns and applying recursive computation.

\subsection{Comparing Alternative Attention Mechanisms}
\label{app:arch}

\begin{figure}[b]
  \centering
  \begin{subfigure}[b]{0.3\textwidth}
    \centering
    \includegraphics[width=\textwidth]{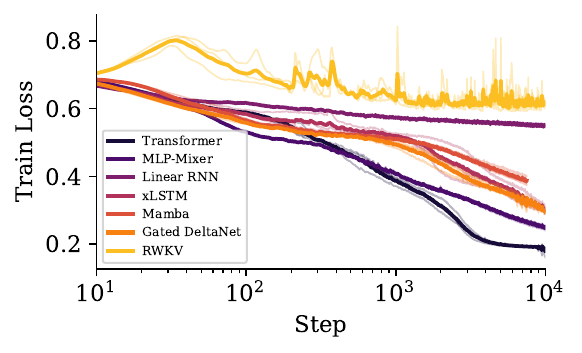}
    \caption{Cellular Automata}
    \label{fig:all_arch_gca}
  \end{subfigure}
  \hfill
  \begin{subfigure}[b]{0.3\textwidth}
    \centering
    \includegraphics[width=\textwidth]{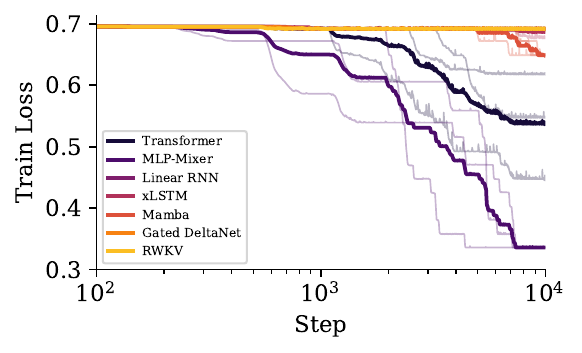}
    \caption{Linear Map ($N=16$)} 
    \label{fig:all_arch_linmap_n16}
  \end{subfigure}
  \hfill
  \begin{subfigure}[b]{0.3\textwidth}
    \centering
    \includegraphics[width=\textwidth]{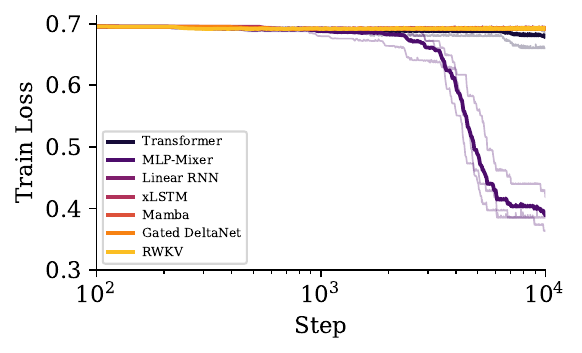}
    \caption{Linear Map ($N=32$)} 
    \label{fig:all_arch_linmap_n32}
  \end{subfigure}
  \caption{\textbf{Alternative architectures underperform relative to transformers on both cellular automata and linear map tasks.} Aside from MLP-Mixer, which only outperforms transformers on the linear map task, all other architectures we evaluate yield subpar performance on both tasks. Notably, as we increase the state size $N$ from $16$ to $32$ for the linear map task, MLP-Mixer still makes reasonable progress during training but all other architectures remain stuck at the initial loss plateau.}
  
  \label{fig:arch_all}
\end{figure}

Transformers model dependencies between tokens through dot-product attention, and in Section~\ref{sec:alt_arch} we compare this mechanism to MLP-Mixer~\citep{tolstikhin2021mlpmixerallmlparchitecturevision}, which learns an MLP over the sequence dimension. In this section, we also evaluate Linear RNN~\citep{orvieto2023resurrectingrecurrentneuralnetworks}, xLSTM~\citep{beck2024xlstmextendedlongshortterm}, Mamba~\citep{gu2024mambalineartimesequencemodeling}, Gated DeltaNet~\citep{yang2025gateddeltanetworksimproving}, and RWKV~\citep{peng2023rwkvreinventingrnnstransformer} as alternatives to the attention-based transformer. Figure~\ref{fig:arch_all} illustrates the results.

As previously demonstrated in  Figure~\ref{fig:arch_linmap}, MLP-Mixer outperforms transformers on the most difficult linear map tasks, but not on the cellular automata task. In Figure~\ref{fig:all_arch_gca} we see that all other architectures also yield worse training performance than a transformer on the cellular automata task. When trained on the linear map task with state size $N=16$ and sparsity $s=3$, only transformer and MLP-Mixer achieve nontrivial performance, and MLP-Mixer learns the task completely. However, when we increase $N$ to $32$, thus extending the context length, all architectures except for MLP-Mixer fail to make any progress at all.

\section{Does Pre-pretraining Accelerate Emergence of Capabilities?}
\label{app:ppt}

Our results in Section~\ref{sec:arch} show how architectural interventions improve a model's ability to learn attention patterns. We now explore how \textit{data} interventions accelerate emergence of practical capabilities in language models. Following prior work, we apply pre-pretraining (PPT) with two synthetic datasets: 33 million tokens of $k$-shuffle Dyck~\citep{hu2025circuitschomskyprepretrainingformal} and 164 million tokens of neural cellular automata~(NCA,~\citet{lee2026traininglanguagemodelsneural}). Our training experiments consist of two main phases. In the PPT phase, we train a randomly initialized model on synthetic data, and in the second phase we train on natural language pretraining data using the final weights from the first phase as an initialization. We compare to a natural language baseline where we \textit{only} train on standard pretraining data from scratch, starting with the same random initialization we would use for PPT.

We train transformer models with 160 million parameters, following the Pythia training recipe exactly~\citep{biderman2023pythiasuiteanalyzinglarge}. We train on 10 billion tokens of C4 data (5,000 steps) since the original dataset used to train Pythia models is no longer available. For pre-pretraining, we follow the implementations from~\citet{hu2025circuitschomskyprepretrainingformal} and~\citet{lee2026traininglanguagemodelsneural} for Dyck and NCA respectively, except we extend the context length of NCA data to 2048 to match that of standard Pythia language training. Note pre-pretraining hyperparameters differ from those used in the natural language training phase. 

\paragraph{Hyperparameters.} Natural language training uses 5,000 steps with 1,430 warmup steps. For Dyck pre-pretraining, we train for 500 steps (30M tokens) with context length 2048 and batch size 32, matching the hyperparameters of~\citet{hu2025circuitschomskyprepretrainingformal} for 160M parameter models. For NCA pre-pretraining, we use the same hyperparameters except with batch size 16, yielding 164M tokens over 5,000 steps. We also re-initialize the token embeddings for NCA pre-pretraining.

\begin{figure}[t]
  \centering
  \begin{subfigure}[b]{0.6\textwidth}
    \centering
    \includegraphics[width=\textwidth]{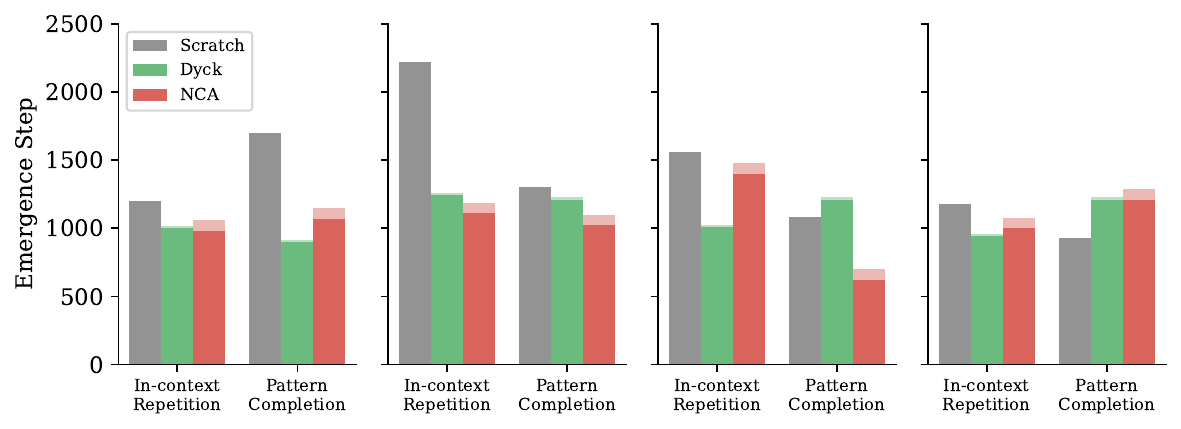}
    \caption{Time to Emergence}
    \label{fig:ppt_emergence}
  \end{subfigure}
  \hfill
  \begin{subfigure}[b]{0.35\textwidth}
    \centering
    \includegraphics[width=\textwidth]{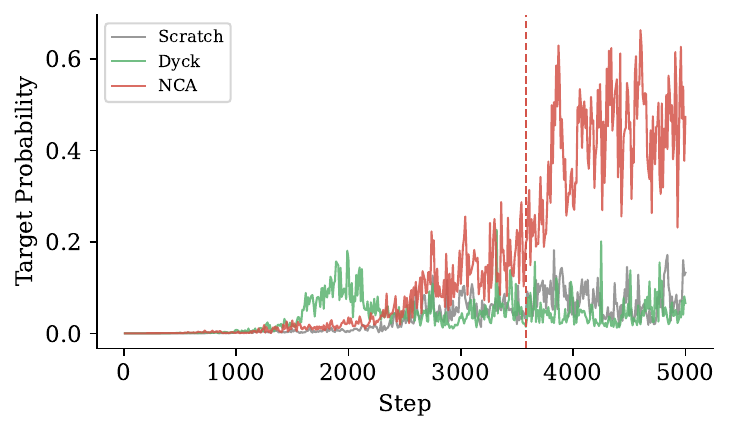}
    \caption{IOI Emergence Across Steps} 
    \label{fig:ppt_ioi}
  \end{subfigure}
  \caption{\textbf{Pre-pretraining improves emergence speed for in-context capabilities.} (a) Prepretraining with Dyck formal languages (green) or neural cellular automata (red) elicits in-context repetition capability earlier than training from scratch (grey) across four random seeds. The uplift persists even when we account for the additional pre-pretraining tokens (shaded bars). However, PPT does not consistently improve emergence for the pattern completion task. (b) For one model seed, NCA pre-pretraining leads to the emergence of indirect object identification (denoted by dashed red line), whereas the same capability does not yet emerge for a model trained from scratch or with Dyck PPT after 5000 steps.}
  \label{fig:ppt}
\end{figure}

We compare training from scratch, PPT with Dyck, and PPT with NCA across 4 different random initialization seeds. As previously demonstrated by~\citet{hu2025circuitschomskyprepretrainingformal} and~\citet{lee2026traininglanguagemodelsneural}, both Dyck and NCA pre-pretraining result in lower final loss after 5,000 steps compared to training from scratch. During training, we probe for the emergence of the capabilities illustrated in Figure~\ref{fig:emergence_tasks} every ten steps. We also track the time to emergence using the same approach described in Section~\ref{sec:emergence_search}.

Figure~\ref{fig:ppt_emergence} shows emergence results for in-context repetition, pattern completion, and indirect object identification. Pre-pretraining \textbf{tends to result in in-context capabilities emerging earlier}, though this trend does not hold for pattern completion. In Figure~\ref{fig:ppt_ioi}, NCA pre-pretraining leads to the emergence of indirect object identification for one seed, whereas the capability does not emerge for training from scratch or Dyck PPT after 5,000 steps. However, we caution that these results are based on only four random seeds, which limits statistical power given the high inter-run variance of emergence. Copying does not emerge in any model after 5,000 steps; we believe this may be due to the change in pretraining dataset from the original Pythia setup. These results suggest that the effect of pre-pretraining on emergent capabilities depends on the specific capability and the nature of the synthetic data, but a larger sample is needed to draw concrete conclusions.

\end{document}